\documentclass[11pt]{article}
\usepackage[preprint]{acl}
\usepackage{times}
\usepackage{latexsym}
\usepackage[T1]{fontenc}
\usepackage[utf8]{inputenc}
\usepackage{microtype}
\usepackage{amsmath}
\usepackage{amssymb}
\usepackage{booktabs}
\usepackage{graphicx}

\title{Temporal Context-Aware Spam Detection Using\\Transformer-Based NLP Models}

\author{Fariz Zeynalov \\
  University of Chicago \\
  \texttt{zeynalov@uchicago.edu} \\\And
  Razim Mammadli \\
  Duke University \\
  \texttt{rm620@duke.edu}}

\begin{document}
\maketitle

\begin{abstract}
Spam detection is usually treated as single-message classification, ignoring behavioral patterns that emerge across sequences of messages from the same sender. We study outbound SMS spam detection at the carrier level, where the goal is to predict at send time whether a carrier will flag a message as spam. Using a proprietary production database of 43,741 outbound SMS messages with UTC transmission timestamps and carrier delivery statuses, we report three sets of results. On an 8,749-message held-out test set, a TF-IDF + logistic regression baseline reaches AUC-ROC 0.8721, and a text-only DeBERTa-v3-base classifier reaches AUC-ROC 0.9011 and F1 0.8412. A deployed DistilBERT model augmented with inter-message temporal features reaches accuracy 0.8238, spam-class F1 0.8382, and recall 0.9128 at threshold 0.5 on a separate, class-balanced 20,000-message production evaluation set; these results are not directly comparable with the held-out test results. We also describe \textit{TemporalBertSpam}, an architecture that encodes each sender's message history with a causal Transformer, and outline the same-split evaluation needed to measure its contribution.
\end{abstract}

\section{Introduction}

SMS carriers run automated filters that flag outbound messages as spam based on content, sender reputation, and behavioral signals. These carrier-side filters differ from classic spam classifiers, which label inbound unsolicited messages: here the task is to predict, at send time, whether the carrier network will block a message. This framing matters in practice, because a blocked message is never delivered and false positives harm legitimate senders.

Most prior NLP work on spam detection classifies single messages with transformer-based text encoders \citep{jamal2024,asliyuksek2025}. These approaches treat each message independently. However, spam campaigns often vary their wording (spintax) to evade per-message text filters while still showing detectable behavior: high-volume bursts, regular intervals between messages, and many recipients reached from a single phone number in a short time.

This paper studies whether information about the temporal sequence of a sender's messages can complement message content. We frame three research questions:

\begin{itemize}
    \item \textbf{RQ1:} Do inter-message temporal features (time delta $\Delta t$ and $\log(1+\Delta t)$) improve detection over a text-only transformer baseline?
    \item \textbf{RQ2:} Does sequence-level modeling of sender histories improve detection over temporal features alone?
    \item \textbf{RQ3:} How do message content type (URL presence, urgency language) and sender volume relate to detection performance?
\end{itemize}

This paper reports baseline results on a held-out test set and production results for a deployed temporal-feature model. Answering RQ1 and RQ2 requires evaluating all systems on a single time-ordered split, and RQ3 requires a subgroup analysis on that same split; we leave both to future work (see Limitations). Our dataset contains per-message timestamps, which are absent from common benchmarks such as the SMS Spam Collection \citep{almeida2011sms}.

\section{Related Work}

\paragraph{Transformer-based text classifiers for spam}
\citet{devlin2019bert} introduced BERT, which became the dominant paradigm for NLP classification. \citet{sanh2019distilbert} distilled BERT into DistilBERT, which is 40\% smaller while retaining 97\% of BERT's language understanding performance, making it well suited to high-throughput inference. \citet{jamal2024} fine-tune DistilBERT and RoBERTa on a three-class phishing/spam/ham task, reporting 97.1\% and 99.0\% test accuracy respectively. Their labels come from human-annotated corpora; our labels reflect network-level carrier decisions on outbound messages.

\paragraph{Combining text with non-textual features}
\citet{asliyuksek2025} combine DistilBERT representations with static structural features of each email and report 99.62\% accuracy on a combined corpus of 81,586 emails from seven datasets. Their non-textual features are computed per message, with no cross-message temporal component. Our approach instead targets behavioral patterns across sequences of messages from the same sender.

\paragraph{Behavioral and temporal signals in spam detection}
Network and security research has long used sender behavior rather than content. \citet{ramachandran2007behavioral} classify email senders by their sending patterns across target domains, and \citet{hao2009snare} detect email spammers with lightweight spatio-temporal network-level features. For SMS, \citet{xu2012sms} detect spammers on the service side using noncontent features, including temporal and network features, without inspecting message text. At carrier scale, \citet{jiang2013understanding} find that spam numbers with similar content show strongly similar sending patterns, and \citet{jiang2013greystar} detect SMS spam numbers from their sending behavior in a large cellular network. Surveys of SMS spam filtering methods and data \citep{delany2012sms} focus mainly on content-based methods. Our work combines a transformer text encoder with temporal features and, in the full TemporalBertSpam design, sequence modeling over each sender's history.

\section{Problem Formulation}

Let a sender $s$ transmit messages $m_1, m_2, \ldots, m_n$ at times $t_1 < t_2 < \cdots < t_n$. Each message $m_i$ has text $x_i$ and a binary carrier label $y_i \in \{0,1\}$, where $y_i = 1$ means the carrier flagged the message as spam and $y_i = 0$ means it was delivered. The task is to predict $y_i$ from the text and temporal context of $m_i$ and its predecessors from the same sender.

\section{Dataset}

\subsection{Production SMS Database}

Our data come from the outbound SMS database of a commercial CRM platform that sends messages through the Telnyx telecommunications API. Records were pulled from the platform's PostgreSQL database in streaming batches of 10,000 rows, joining message records (text, sender number, send timestamp) with carrier delivery metadata (status and error codes).

\paragraph{Data scope} The raw pull contains 59,342 outbound SMS messages sent between September 22 and December 19, 2025 (about 89 days), from multiple outbound phone numbers used by CRM customers for appointment reminders, marketing, and transactional notifications. We keep only the two terminal statuses that reflect carrier filtering decisions, \texttt{spam} (carrier-flagged) and \texttt{delivered}, leaving 43,741 messages. Other statuses (\texttt{delivery\_\allowbreak{}failed}, \texttt{queued}, \texttt{blocked}, \texttt{error}, \texttt{delivery\_\allowbreak{}unconfirmed}, \texttt{sending\_\allowbreak{}failed}, \texttt{sent}, \texttt{read}) do not cleanly reflect carrier content filtering and are excluded.

\paragraph{Label semantics} Labels come from carrier delivery statuses, not human annotation. A message labeled \texttt{spam} ($y=1$) was flagged by the carrier network as unsolicited or policy-violating; a message labeled \texttt{delivered} ($y=0$) reached the recipient device.

\paragraph{Class distribution} The dataset contains 18,283 spam messages (41.8\%) and 25,458 delivered messages (58.2\%). We report AUC-ROC alongside F1 because of this imbalance.

\paragraph{Timestamps and sender identity} Each message has a UTC transmission timestamp at microsecond precision, recorded from Telnyx webhook events. The sender field holds the outbound phone number (E.164 format), which we use only to group messages into per-sender sequences. Sender numbers are not reported anywhere in this paper.

\subsection{Preprocessing}

\paragraph{Near-duplicate templates} Spam campaigns often send many near-duplicate variants of one template. No messages were removed for near-duplication, so all 43,741 messages are used for modeling and near-duplicate variants may remain in the data. Section~\ref{sec:dedup} describes a screening procedure for future experiments.

\paragraph{Time normalization} Unix timestamps (seconds) are min-max normalized over the corpus:
\[
t_{\text{norm}} = \frac{t - t_{\min}}{t_{\max} - t_{\min}}.
\]

\paragraph{Train/test split} For the text baselines we use an 80/20 split stratified on the label, giving 34,992 training and 8,749 test messages (41.8\% spam in each). Because this split is random, messages from the same sender and template can appear in both partitions. For the sequence model we specify a time-ordered split, with the earliest messages used for training and the most recent for testing, to avoid temporal leakage.

\section{Methods}

\subsection{TF-IDF + Logistic Regression}

The weakest baseline uses TF-IDF features (unigrams and bigrams, top 50,000 features, sublinear TF) with L2-regularized logistic regression ($C = 1.0$, \texttt{liblinear} solver).

\subsection{Text-Only DeBERTa-v3-base}
\label{sec:deberta}

Our main text-only baseline fine-tunes DeBERTa-v3-base \citep{he2023debertav3} (86M backbone parameters) for binary classification with \texttt{BCEWith\-Logits\-Loss} and a single output logit. DeBERTa uses disentangled attention over content and position representations \citep{he2021deberta}, and DeBERTaV3 adds ELECTRA-style pre-training \citep{he2023debertav3}. We use it as a strong text-only reference to test whether temporal information adds value beyond a strong content encoder.

\textbf{Training details:} AdamW ($\text{lr} = 2 \times 10^{-5}$, weight decay 0), batch size 32, 3 epochs, dynamic padding, maximum sequence length 128 tokens.

\textbf{Label convention:} In this experiment delivered messages are the positive class (label 1) and spam messages the negative class (label 0), following the platform's internal convention. AUC-ROC is unaffected by this choice. The reported F1 is for the delivered class.

\subsection{DistilBERT + Temporal Features}

This model augments DistilBERT's CLS representation with two temporal features: the normalized time delta $\Delta t_i = t_i - t_{i-1}$ (0 for a sender's first message) and $\log(1+\Delta t_i)$. The features are projected to the CLS dimension by a learned linear layer and added to the CLS vector before the classification head. It uses temporal features without sequence modeling, and it is the model deployed in production.

\subsection{TemporalBertSpam}

The full model treats each sender's history as a causal sequence and applies a second-stage Transformer over message representations.

\paragraph{Stage 1: Text encoding} Each message $x_i$ is encoded independently with \texttt{distilbert-\allowbreak base-\allowbreak uncased}; the final-layer CLS vector gives a 768-dimensional representation $h_i$.

\paragraph{Stage 2: Temporal feature projection} The two temporal features are projected to 768 dimensions by a learned matrix $W_{\text{time}} \in \mathbb{R}^{768 \times 2}$ and added to $h_i$:
\[
\tilde{h}_i = h_i + W_{\text{time}} \, [\Delta t_i,\; \log(1+\Delta t_i)]^\top.
\]

\paragraph{Stage 3: Sequence-level positional encoding} The representations $\tilde{h}_1, \ldots, \tilde{h}_T$ in a window receive sinusoidal positional encodings \citep{vaswani2017attention}:
\begin{align*}
\text{PE}(p,2k) &= \sin\!\left(p / 10000^{2k/d}\right),\\
\text{PE}(p,2k+1) &= \cos\!\left(p / 10000^{2k/d}\right),
\end{align*}
where $p \in \{0,\ldots,T-1\}$ is the position of the message within the window and $d = 768$. These encodings are distinct from DistilBERT's learned token-position embeddings: DistilBERT's embeddings encode the position of each token within one message, while the sinusoidal encoding marks the position of each message within the sender's history, inside a separate Transformer.

\paragraph{Stage 4: Causal Transformer encoder} The sequence passes through a 4-layer Transformer encoder \citep{vaswani2017attention} with 8 heads, feedforward dimension 2048, pre-layer normalization, GELU, and dropout 0.1. An upper-triangular attention mask makes position $i$ attend only to positions $0, \ldots, i$. A 2-class linear head produces per-message logits.

\paragraph{Sequence construction} Messages are grouped by sender, sorted by time, and split into sliding windows of $T=32$ (stride 16 for training, 32 for evaluation). Shorter histories are left-padded with a padding mask. For a sender's first message, no earlier context exists, so the prediction depends on that message's content and temporal features only.

\textbf{Training configuration.} AdamW ($\text{lr} = 2\times10^{-5}$, weight decay 0.01), batch size 8, 5 epochs, linear warmup over 10\% of steps, mixed precision, gradient clipping at norm 1.0, and cross-entropy over per-message predictions in each window.

\section{Evaluation Setup}

\paragraph{Metrics} We report AUC-ROC (threshold-free and invariant to label encoding) for the text baselines, and F1 and accuracy for all models. Because the class balance differs between our two evaluation sets, we state the positive class for each F1 value.

\paragraph{Evaluation sets} Text baselines are evaluated on the 8,749-message held-out test set. The deployed DistilBERT + temporal model is evaluated on a 20,000-message set drawn from the production monitoring system, with 10,000 spam and 10,000 delivered messages. Results across the two sets are not directly comparable.

\section{Results}

\subsection{Held-Out Test Set}

\begin{table}[t]
\centering
\small
\begin{tabular}{lccc}
\toprule
\textbf{Model} & \textbf{AUC-ROC} & \textbf{F1} & \textbf{Acc.} \\
\midrule
TF-IDF + LR     & 0.8721 & --     & 0.7912 \\
DeBERTa-v3-base & 0.9011 & 0.8412 & 0.8340 \\
\bottomrule
\end{tabular}
\caption{Text baselines on the 8,749-message held-out test set (80/20 stratified split). F1 is for the delivered class (platform label convention, Section~\ref{sec:deberta}) and is reported for DeBERTa only.}
\label{tab:heldout}
\end{table}

Table~\ref{tab:heldout} shows the text baselines. DeBERTa reaches AUC-ROC 0.9011 and F1 0.8412, so the task is largely learnable from content alone, and the stronger encoder improves AUC by about 3 points over TF-IDF. These numbers are far below the 97--99\% accuracy reported by \citet{jamal2024}, but the tasks differ in domain, label source, and class setup, so the results are not directly comparable. Content-only models cannot see a sender's history: two messages with identical text can carry different risk depending on whether they are a sender's first message or one of hundreds sent in a burst.

\subsection{Production Deployment Results}

\begin{table}[t]
\centering
\small
\begin{tabular}{lc}
\toprule
\textbf{Metric} & \textbf{Value} \\
\midrule
Accuracy            & 0.8238 \\
Precision (spam)    & 0.7749 \\
Recall (spam)       & 0.9128 \\
F1 (spam)           & 0.8382 \\
F1 (delivered)      & 0.8066 \\
Macro F1            & 0.8224 \\
\midrule
True positives      & 9,128 \\
False positives     & 2,652 \\
True negatives      & 7,348 \\
False negatives     & 872 \\
Avg.\ confidence    & 0.753 \\
\bottomrule
\end{tabular}
\caption{Deployed DistilBERT + temporal model on the 20,000-message production evaluation set (10,000 spam, 10,000 delivered) at threshold 0.5. Spam is the positive class.}
\label{tab:production}
\end{table}

Table~\ref{tab:production} reports the deployed model at threshold 0.5. The model favors recall: it catches 91.3\% of carrier-flagged spam, at the cost of 2,652 false positives among 10,000 delivered messages.

\begin{table}[t]
\centering
\small
\setlength{\tabcolsep}{4pt}
\begin{tabular}{ccccccc}
\toprule
\textbf{Thr.} & \textbf{Acc.} & \textbf{Prec.} & \textbf{Rec.} & \textbf{F1} & \textbf{FP} & \textbf{FN} \\
\midrule
0.30 & 0.814 & 0.757 & 0.926 & 0.833 & 2972 & 744 \\
0.35 & 0.815 & 0.760 & 0.920 & 0.833 & 2900 & 796 \\
0.40 & 0.817 & 0.764 & 0.918 & 0.834 & 2840 & 816 \\
0.45 & 0.821 & 0.769 & 0.916 & 0.836 & 2752 & 836 \\
0.50 & 0.824 & 0.775 & 0.913 & 0.838 & 2652 & 872 \\
\textbf{0.55} & \textbf{0.825} & \textbf{0.779} & \textbf{0.909} & \textbf{0.839} & \textbf{2584} & \textbf{912} \\
0.60 & 0.825 & 0.782 & 0.902 & 0.838 & 2512 & 980 \\
0.65 & 0.826 & 0.785 & 0.899 & 0.838 & 2468 & 1012 \\
0.70 & 0.824 & 0.786 & 0.888 & 0.834 & 2412 & 1116 \\
0.75 & 0.807 & 0.801 & 0.816 & 0.808 & 2032 & 1836 \\
0.80 & 0.734 & 0.848 & 0.571 & 0.682 & 1024 & 4292 \\
\bottomrule
\end{tabular}
\caption{Threshold sweep on the production evaluation set. Best F1 (0.839) is at threshold 0.55 (bold). FP = false positives; FN = false negatives.}
\label{tab:threshold}
\end{table}

\begin{figure}[t]
  \centering
  \includegraphics[width=\columnwidth]{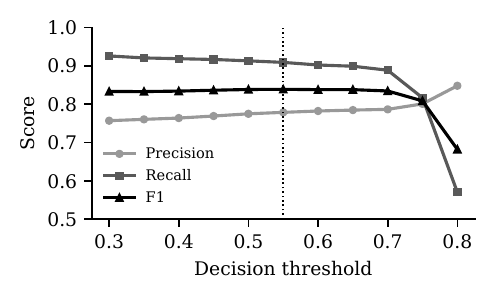}
  \caption{Precision, recall, and F1 across decision thresholds on the production evaluation set (from Table~\ref{tab:threshold}). The dotted line marks the production threshold of 0.55.}
  \label{fig:threshold}
\end{figure}

\paragraph{Threshold calibration} Table~\ref{tab:threshold} and Figure~\ref{fig:threshold} show performance for thresholds from 0.30 to 0.80. F1 is nearly flat between 0.45 and 0.65 and peaks at 0.55 (0.839), a marginal gain over 0.5 (0.838). Recall falls sharply above 0.70. Production uses a threshold of 0.55. Because the threshold was chosen on the same set used for evaluation, these figures are optimistic.

\begin{figure}[t]
  \centering
  \includegraphics[width=\columnwidth]{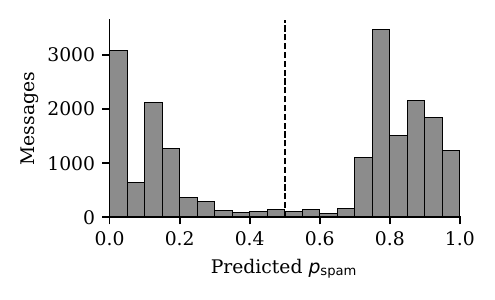}
  \caption{Distribution of predicted $p_{\text{spam}}$ on the 20,000-message production evaluation set (0.05-wide bins). The dashed line marks threshold 0.5. Most predictions fall below 0.20 or above 0.70, with few in between.}
  \label{fig:hist}
\end{figure}

\paragraph{Score distribution and calibration} Figure~\ref{fig:hist} shows that predictions are bimodal: 7,112 messages score below 0.20 and 11,296 score at or above 0.70, with only 1,592 in between. Table~\ref{tab:calibration} reports calibration in 0.1-wide probability buckets below 0.6. The two lowest buckets are well calibrated (error at most 0.011). Bucket 2 underestimates spam (predicted 0.247, observed 0.339), and buckets 4 and 5 overestimate it, with errors of 0.227 and 0.116. Predictions at or above 0.6 (11,532 messages) have an observed spam rate of 0.782, equal to the precision at threshold 0.6 in Table~\ref{tab:threshold}.

\begin{table}[t]
\centering
\small
\begin{tabular}{ccccc}
\toprule
\textbf{Bucket} & \textbf{n} & \textbf{Avg.\ pred.} & \textbf{Observed} & \textbf{Error} \\
\midrule
$[0.0, 0.1)$ & 3732 & 0.025 & 0.021 & 0.004 \\
$[0.1, 0.2)$ & 3380 & 0.141 & 0.130 & 0.011 \\
$[0.2, 0.3)$ & 660  & 0.247 & 0.339 & 0.093 \\
$[0.3, 0.4)$ & 204  & 0.338 & 0.353 & 0.015 \\
$[0.4, 0.5)$ & 244  & 0.456 & 0.230 & 0.227 \\
$[0.5, 0.6)$ & 248  & 0.552 & 0.436 & 0.116 \\
\bottomrule
\end{tabular}
\caption{Calibration for predictions below 0.6 (8,468 of 20,000 messages). Observed is the actual spam rate in each bucket; Error is the absolute difference from the average prediction.}
\label{tab:calibration}
\end{table}

\paragraph{Model version history} Table~\ref{tab:versions} lists the three versions promoted to production. Accuracy rose by 10.3 points between the first and second versions, while recall fell from 0.936 to 0.916, indicating a shift toward fewer false positives.

\begin{table}[t]
\centering
\small
\setlength{\tabcolsep}{4pt}
\begin{tabular}{lcccc}
\toprule
\textbf{Version} & \textbf{Val.\ rows} & \textbf{Acc.} & \textbf{Prec.} & \textbf{Rec.} \\
\midrule
v20260502\_223227 & 15k & 0.717 & 0.651 & 0.936 \\
v20260506\_162645 & 15k & 0.820 & 0.768 & 0.916 \\
v20260506\_223403 & 20k & 0.824 & 0.775 & 0.913 \\
\bottomrule
\end{tabular}
\caption{Production model version history.}
\label{tab:versions}
\end{table}

\section{Discussion}

\subsection{Expected Behavior of the Full Model}

The following are design hypotheses for TemporalBertSpam, not measured results.

\paragraph{High-volume campaigns} Burst sending produces short intervals ($\Delta t_i \approx 0$) and many messages from one sender. The sequence encoder is designed to associate such patterns with higher spam probability, independently of text, which a per-message classifier cannot observe.

\paragraph{Spintax evasion} When individual variants look benign to the text encoder, the sending pattern (regular intervals, many messages from one number) may still provide a useful signal through the causal attention over earlier messages.

\paragraph{Cold start} For a sender's first message, no history exists, and the model relies on content and the temporal features of that message alone.

\subsection{Near-Duplicate Screening for Future Experiments}
\label{sec:dedup}

We propose the following screening procedure. Each message is lowercased, whitespace is collapsed, and the text is represented as a TF-IDF vector over unigrams and bigrams, with weight $\text{tf}'(t,d) \cdot \log(N/\text{df}(t))$ and sublinear term frequency $\text{tf}' = 1 + \log(\text{tf})$. Vectors are L2-normalized, so cosine similarity reduces to a dot product. Messages are processed in corpus order; a message is discarded if its maximum cosine similarity to any retained message is at least a threshold such as 0.9.

A high threshold catches variants that differ in only a few tokens, but spintax variants that swap many words or phrases share less vocabulary and fall below it. Since near-duplicates across the train/test split can inflate metrics, combining lexical screening with embedding-based deduplication \citep{reimers2019sbert} is a priority for future experiments.

\subsection{Positional Encoding Design}

An alternative design would feed temporal features as extra token embeddings into DistilBERT. That would mix inter-message information into a model pre-trained only on intra-message token sequences. Keeping a separate sequence Transformer with its own sinusoidal encoding keeps the two levels apart and makes the ablation clean: removing the sequence Transformer (the DistilBERT + temporal model) changes nothing in the text encoder. Sinusoidal encoding also adds no parameters, which matters given the limited number of distinct sender histories.

\subsection{Comparison to \citet{asliyuksek2025}}

\citet{asliyuksek2025} combine DistilBERT with static structural features and reach 99.62\% accuracy on a combined email corpus. Our setting differs in three ways: (1) labels come from carrier decisions rather than curated corpora, which adds noise but reflects real network behavior; (2) our non-textual features are cross-message temporal signals rather than per-message statistics; (3) TemporalBertSpam explicitly models dependencies across a sender's messages. Given these differences in domain, labels, and data, accuracy figures are not directly comparable.

\subsection{Newer Encoders}

DistilBERT (2019) and DeBERTaV3 (2021) predate several stronger encoders. ModernBERT \citep{warner2025modernbert} is trained on 2 trillion tokens with a native 8,192-token context; for SMS, which is short, its benefit would come mainly from better pre-training rather than longer context. Sentence-BERT \citep{reimers2019sbert} and later embedding models could replace TF-IDF deduplication with semantic deduplication that catches paraphrased spintax. LLM-based embeddings could better capture commercial intent in novel phrasings.

A stronger text encoder would mainly help with hard cases: messages that are spam in intent but phrased in ways the encoder's pre-training data did not cover. Behavioral signals such as burst sending and short intervals are independent of content, so we expect temporal information to remain useful even with a stronger encoder; testing this with ModernBERT or an LLM encoder at Stage 1 is a natural next experiment.

\section{Conclusion}

We study outbound SMS spam detection with carrier delivery labels and real transmission timestamps, using 43,741 production messages. A text-only DeBERTa-v3-base baseline reaches AUC-ROC 0.9011 and F1 0.8412 on a held-out test set, and a deployed DistilBERT + temporal model reaches spam-class F1 0.8382 and recall 0.9128 at threshold 0.5 on a separate, balanced 20,000-message production set. Unlike common public SMS spam benchmarks, the dataset includes per-message timestamps and carrier-decision labels. We also describe TemporalBertSpam, which models each sender's history with a causal Transformer.

The key next step is to evaluate the TF-IDF, DeBERTa, DistilBERT + temporal, and TemporalBertSpam models on one time-ordered split, which would directly answer RQ1 and RQ2, together with a subgroup analysis by message type and sender volume on the same split (RQ3). Further directions include near-duplicate removal, newer encoders at Stage 1, and few-shot methods for senders with sparse histories.

\section*{Limitations}

\paragraph{Incomplete comparison} The text baselines and the deployed model are evaluated on different sets with different class balance (41.8\% vs.\ 50\% spam), so we cannot yet attribute performance differences to temporal features. The full TemporalBertSpam model has not been evaluated in this paper.

\paragraph{Evaluation protocol} The text baselines use a random split, so messages from the same sender or template can appear in both training and test data, which may inflate their scores. The production evaluation set was drawn from the monitoring system, and we do not document its sampling procedure or rule out overlap with the validation data of the deployed model version. The production decision threshold was selected on the same set used for evaluation. Production results should therefore be read as optimistic.

\paragraph{Label noise and selection bias} Carrier statuses are an imperfect proxy for spam, and carrier filtering policies may change over time. The production system uses a stochastic send policy that holds some messages; held messages receive no carrier label, so the labeled data are not a random sample of all attempted messages.

\paragraph{Near-duplicates} No near-duplicate removal was applied, so templated variants may appear in both training and test data.

\paragraph{Generality} The data come from one CRM platform, one messaging provider, and an 89-day window, and the dataset cannot be released because it is proprietary. Results may not transfer to other platforms, regions, or periods.

\section*{Ethics Statement}

Message text contains real SMS content sent by CRM users. We report only aggregate statistics and model metrics; and no message text, phone number, or sender identifier is reproduced. The data were accessed under the data use agreement governing the authors' access to the platform's systems.

\bibliography{references}

\end{document}